\documentclass[runningheads]{llncs}

\usepackage{eccv}

\usepackage{eccvabbrv}

\usepackage{graphicx}
\usepackage{multirow}
\usepackage{multicol}
\usepackage{caption}
\usepackage{booktabs}
\usepackage{tabularx}
\newcolumntype{Y}{>{\centering\arraybackslash}X}
\usepackage{hyperref}
\usepackage{threeparttable}
\usepackage{bbding}
\usepackage{wrapfig}

\usepackage{amssymb}
\makeatletter

\newcommand{\Rmnum}[1]{\expandafter\@slowromancap\romannumeral #1@}
\makeatother

\usepackage[accsupp]{axessibility}  
\usepackage{pdfpages}
\usepackage[misc]{ifsym}
\usepackage{pifont}
\newcommand{\cmark}{\ding{51}}%
\newcommand{\xmark}{\ding{55}}%

\usepackage{hyperref}

\usepackage{orcidlink}

\begin{document}

\title{Toward Open Vocabulary Aerial Object Detection with CLIP-Activated Student-Teacher Learning} 

\titlerunning{CastDet}

\author{Yan Li\inst{1}\and
Weiwei Guo\inst{2\text{ \Letter}}\and
Xue Yang\inst{3}\and
Ning Liao\inst{1} \\
Dunyun He\inst{2}\and
Jiaqi Zhou\inst{1}\and
Wenxian Yu\inst{1\text{ \Letter}}
}

\authorrunning{Y. Li et al.}

\institute{$^1$Shanghai Jiao Tong University \hspace{3pt} $^2$Tongji University \\ $^3$OpenGVLab, Shanghai AI Laboratory}

\maketitle

\begin{abstract}
 An increasingly massive number of remote-sensing images spurs the development of extensible object detectors that can detect objects beyond training categories without costly collecting new labeled data. In this paper, we aim to develop open-vocabulary object detection (OVD) technique in aerial images that scales up object vocabulary size beyond training data. The performance of OVD greatly relies on the quality of class-agnostic region proposals and pseudo-labels for novel object categories. To simultaneously generate high-quality proposals and pseudo-labels, we propose \textbf{CastDet}, a \textbf{C}LIP-\textbf{a}ctivated \textbf{s}tudent-\textbf{t}eacher open-vocabulary object \textbf{Det}ection framework.  Our end-to-end framework following the student-teacher self-learning mechanism employs the RemoteCLIP model as an extra omniscient teacher with rich knowledge. By doing so, our approach boosts not only novel object proposals but also classification. Furthermore, we devise a dynamic label queue strategy to maintain high-quality pseudo labels during batch training. We conduct extensive experiments on multiple existing aerial object detection datasets, which are set up for the OVD task. Experimental results demonstrate our CastDet achieving superior open-vocabulary detection performance, e.g., reaching 46.5\% mAP  on VisDroneZSD novel categories, which outperforms the state-of-the-art open-vocabulary detectors by 21.0\% mAP. To our best knowledge, this is the first work to apply and develop the open-vocabulary object detection technique for aerial images. The code is available at \href{https://github.com/lizzy8587/CastDet}{https://github.com/lizzy8587/CastDet}.
\end{abstract}

\section{Introduction}
Object detection in aerial images refers to localizing objects of interest on the surface of the earth and predicting their categories which is a pivotal remote sensing image interpretation task for various earth observation applications such as urban management, environmental monitoring and disaster 
search and rescue \cite{zhao2003car, reilly2010detection,sadgrove2018real}. While numerous aerial object detectors have been developed with the adoption of deep learning~\cite{7926624,ding2019learning,yang2021r3det,qian2022rsdet++,yang2019scrdet,yang2019clustered}, they fail to detect objects beyond the training categories. A conventional idea to expand the detectors to novel categories is collecting and annotating large-scale aerial images of rich object categories, which is quite challenging for remote sensing images. This paper advocates more flexible object detectors that can detect novel object categories unseen during the training process to overcome the limitation, currently known as open vocabulary object detection (OVD). It enables us to characterize new objects that emerged in the earth observation data without extra annotation data in open scenarios.  

\begin{wrapfigure}{r}{0.48\textwidth}
    \centering
    \includegraphics[width=1.0\linewidth]{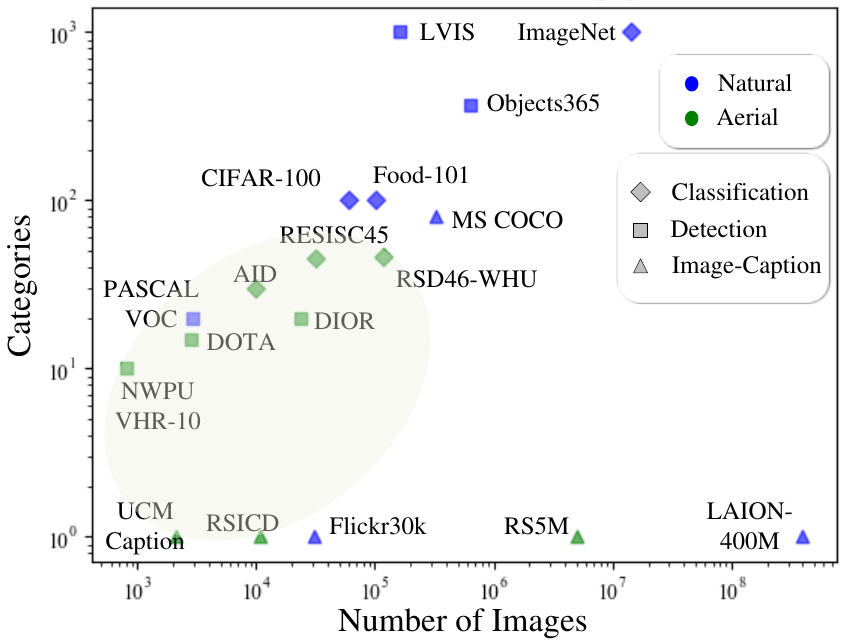}
    \caption{Comparison of target categories and the number of images for 18 common aerial and natural image datasets. \textbf{Challenge 1: Aerial datasets are much smaller} in size and category vocabularies than nature image datasets.}
    \label{fig:dataset_scale}
\end{wrapfigure}

Drawing inspiration from the recent success of OVD in natural images \cite{9879567glip, yao2022detclip, liu2023groundingdino}, we intend to explore challenging open vocabulary object detection for aerial images taken from overhead viewpoints, where the objects exhibit a broader range of variations in scales, orientations, and weak feature appearance\cite{Zhang2023}. In addition, sufficient and accurate annotations for the detector training are time- and labor-expensive, even requiring human experts to curate the datasets. It hinders the detector scaling up in open-world scenarios. As a result, current aerial object detection datasets~\cite{nwpu_cheng2016learning, dior_li2020object, DOTA8578516, levir_zou2017random, visdrone_zhu2021detection}, despite extensive collection efforts, are smaller in size and category vocabularies compared to natural image datasets~\cite{lin2014microsoft,gupta2019lvis,5206848imagenet}. For instance, the existing remote sensing object detection datasets only encompass around 20 categories, much less than the real number of object categories on our earth surface, whereas natural image datasets span thousands of categories, as depicted in Fig.~\ref{fig:dataset_scale}. Their sizes are also relatively small, compared to the natural image datasets. These factors, on the one hand, spur us to develop extensible aerial image object detectors covering more object classes without extra annotation; and, on the other hand, pose challenges to directly applying current OVD methods for natural images to aerial images.

\begin{figure}[t]
  \centering
\includegraphics[width=\linewidth]{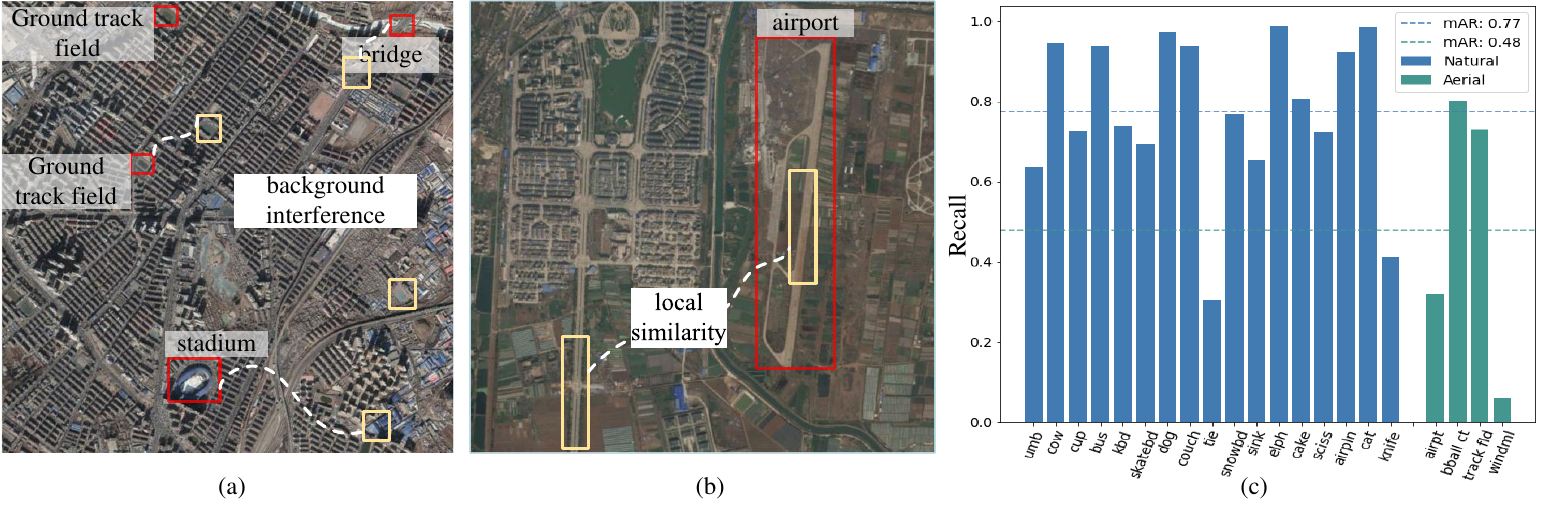}
   \caption{\textbf{Challenge 2: The recall of aerial images is much lower than that of natural images.} (a)(b) Aerial images from DIOR~\cite{dior_li2020object}.  Objects in aerial images exhibit background interference. (c) Class-agnostic RPN recall statistics of novel categories in natural dataset COCO~\cite{lin2014microsoft} and aerial dataset VisDroneZSD~\cite{VisDrone2023} (i.e., 77\% v.s. 48\%).}
   \label{fig:dataset_compare}
\end{figure}

The natural images taken from front viewpoints often exhibit clear contours and texture for which a class-agnostic region proposal network (RPN) trained on a wealthy number of object categories shows excellent generalization capability of proposal generation for unseen categories~\cite{chen2023ovarnet, zhou2022detic}. In contrast, aerial images taken from an overhead perspective can only capture weak appearance features on the top surface of the objects. It often occurs that the objects interfere with the surrounding background with similar appearances, complicating the discrimination between the objects of interest and background noise. For example, AIRPORT is locally similar to HIGHWAY, and common datasets often consider HIGHWAY as background, making it difficult for the model to detect the novel category AIRPORT, as illustrated in Fig.~\ref{fig:dataset_compare}(a)$\sim$(b). It degrades recall for novel categories within aerial imagery OVD, as shown in Fig.~\ref{fig:dataset_compare}(c). To develop an open-vocabulary detector without the expensive annotation, the core question we should ask is: \textit{How to improve OVD precision and recall in aerial images with limited labeled data}?

To address the above issues, we present a simple but effective aerial open vocabulary object detection framework, \textbf{CastDet}, a \textbf{C}LIP-\textbf{a}ctivated \textbf{s}tudent-\textbf{t}eacher \textbf{det}ector. Our aerial OVD detection framework follows the multi-teacher self-learning mechanism, comprising three models: a student model responsible for the detector training, which is guided by two teacher models, a localization teacher model mainly for discovering and localizing potential objects, and an external teacher for classifying novel categories as extra pseudo-labels. Student-teacher learning paradigm is a powerful knowledge distillation and learning framework that commonly comprises a teacher model and a student network learning guided by the teacher model for various learning and vision tasks, including semi-supervised object detection\cite{wang2021knowledge,softteacher_xu2021end}. However, they only work in a closed-set setting but are incapable of discovering and recognizing novel object categories~\cite{tarvainen2017mean,softteacher_xu2021end} not encountered in the training data. To tackle this problem, we incorporate RemoteCLIP~\cite{liu2023remoteclip} as the extra teacher with rich external knowledge into the student-teacher learning process. The RemoteCLIP is a vision-language fundamental model for remote sensing image interpretation pre-trained on large-scale remote sensing image-text pairs following the CLIP\cite{CLIP_radford2021learning}, yielding remarkable generalization ability. Furthermore, in order to maintain the high quality of pseudo-labels and knowledge distillation for the ``unseen'' objects during the model batch training, we propose a dynamic label queue to store and iteratively update the pseudo labels obtained from RemoteCLIP. A hybrid training regime is proposed with the labeled data with the ground truth and the unlabeled data with the pseudo-labels generated by the localization teacher as well as the pseudo-labeled data in the dynamic label queue by the external teacher.

Unlike previous CLIP-based approaches~\cite{gu2021open,zhong2022regionclip,chen2023ovarnet} that directly transfer knowledge from CLIP for zero-shot recognition, our CLIP-activated student-teacher interactive self-learning framework incorporates high-confidence knowledge from RemoteCLIP as an incentive to guide the student and localization teacher to update their knowledge base. Our interactive self-learning mechanism facilitates a ``flywheel effect'' wherein the external teacher transfers knowledge to strengthen the localization teacher to discover potential regions of the ``unseen'' objects and identify their classes while the localization teacher, in turn, generates more accurate pseudo boxes for the external teacher to obtain more accurate pseudo-labels. Through our student-teacher interactive learning scheme, our detection model can be progressively updated to localize and recognize continuously expanded object category vocabulary, improving recall and accuracy.

To the best of our knowledge, this is the first work to address OVD for aerial images, and few benchmark datasets are available. We conduct extensive experiments to evaluate our method by comparing with the state-of-the-art open-vocabulary detectors. Additionally, we assess its performance on several existing aerial object detection datasets.  We split the base and novel categories on these datasets, following the dataset setting in the zero-shot object detection challenge of VisDrone2023\cite{VisDrone2023}. Our method achieves 46.5\% mAP$_{\mathrm{novel}}$, surpassing  21.0\% mAP$_{\mathrm{novel}}$ on VisDroneZSD, showing the superiority of our proposed approach. Our contributions in this paper are summarized as follows:

\begin{enumerate}
    \item Our work pioneers open vocabulary object detection in aerial imagery, aiming at addressing the fundamental challenges in earth observation image interpretation.
    \item We propose CastDet, a novel flexible open-vocabulary detection framework incorporating a student-(multi)teacher self-learning paradigm. Through our hybrid training strategy, the detection model progressively expands the object vocabulary without extra annotation efforts.
    \item We propose a dynamic label queue for storing and incrementally updating high-quality pseudo-labels, enabling us to maintain richer and more accurate labels dynamically.
    \item We utilize several public aerial image datasets to set the open-vocabulary aerial detection benchmark and perform extensive experiments to validate our methods, achieving promising performance improvement.
\end{enumerate}

\section{Related Work}
\textbf{Aerial Image Object Detection}
aims to predict the bounding box coordinates and their corresponding categories in aerial images. Inspired by the remarkable success of deep learning‐based object detection methods for natural images, many researchers have adopted the object detection frameworks originally developed for natural images to aerial images, e.g., Faster R-CNN~\cite{ren2015faster}, RetinaNet~\cite{retina_lin2017focal}, YOLO~\cite{yolo_redmon2016you}, DETR~\cite{carion2020end}, etc\cite{Zhang2023} and tackle peculiar challenges in aerial image object detection, including significant variations in object orientation, scale, and dense object clusters, et al.  To provide a more accurate representation for irregularly shaped or oriented aerial objects, recent work turns to rotated object detection, introducing rotated bounding boxes to align with the object's orientation, e.g., ROI-Transformer~\cite{ding2019learning}, R$^3$Det~\cite{yang2021r3det}, RSDet++~\cite{qian2022rsdet++}. Furthermore, another line of work has concentrated on tackling the challenge of tiny and dense object detection, e.g., SCRDet~\cite{yang2019scrdet}, ClusDet~\cite{yang2019clustered}. Although these aerial object detectors can address specific challenges inherent in aerial image object detection, all of them are trained and evaluated on a pre-defined set of object categories, i.e., closet-set setting, which remains the same during training and testing.  To expand the detector for novel categories absent in training data, we have to re-collect enough labeled training data for novel categories, which is very labor- and time-intensive. In this paper, we intend to develop the first open vocabulary object detector for aerial images to overcome this limitation.

\noindent
\textbf{Open-vocabulary Object Detection}
aims to detect objects beyond the training categories. OVR-CNN\cite{zareian2021open} introduces the inaugural approach to OVD, using bounding box annotations for a limited set of categories as well as a corpus of image-caption pairs to acquire an unbounded vocabulary of concepts. 
Thanks to the remarkable zero-shot transferring capabilities of the pre-trained Vision-Language Models (VLM), e.g., CLIP\cite{CLIP_radford2021learning} and ALIGN\cite{Align_jia2021scaling}, recent OVD methods transfer knowledge from pre-trained VLMs with prompt learning~\cite{du2022learning, feng2022promptdet,zang2022open} or region-level fine-tuning~\cite{gu2021open,zhong2022regionclip,zhou2022detic,wu2023cora} to achieve flexible and versatile detection of extensible object categories. ViLD\cite{gu2021open} transfers knowledge from a pre-trained VLM to a two-stage detector via vision and language knowledge distillation. RegionCLIP\cite{zhong2022regionclip} aligns region-level visual representations with textual concepts. Detic\cite{zhou2022detic} enhances detector vocabulary by training classifiers on image classification data, broadening the range of detectable concepts to tens of thousands. PromptDet\cite{feng2022promptdet} and DetPro~\cite{du2022learning} carefully design the prompt embeddings to better align with the region features. The success of these approaches relies on the following conditions: 1) well-generalized object proposal generation outside training object categories; and 2) large-scale image-text datasets for training to gain the ability of zero-shot classification. Due to the relatively small scale of the existing aerial image object detection dataset and the intrinsic appearance distinction compared to the natural images causing the low recall of region proposal generation for extensive object categories, the OVD methods for natural images can not be directly applied for aerial images, achieving satisfactory performance. 

\section{Methodology}
In this section, we firstly describe our problem setting (Sec. \ref{subsec:preliminaries}), followed by an overview of our CastDet framework (Sec. \ref{subsec:ovd}), and then we introduce the localization teacher as well as the reliable pseudo-bounding-boxes selection strategy (Sec. \ref{sec:localization_teacher}). Finally, we elaborate on the proposed dynamic pseudo label queue to maintain the training samples with high-confidence pseudo-labels  (Sec. \ref{sec:dynamic_queue}) and the hybrid training strategy (Sec. \ref{sec:hybrid_training}). 
\subsection{Problem Description}
\label{subsec:preliminaries}
Given a labeled detection dataset $\mathcal{L}$ with annotations on a set of base categories $\mathcal{C}_{\text{base}}$, and an unlabeled dataset $\mathcal{U}$ that may contain novel categories $\mathcal{C}_{\text{novel}}$. Our training dataset includes both labeled data and unlabled data, i.e., $\mathcal{D}_{\text{train}}=\mathcal{L}\ \cup\ \mathcal{U}=\{(I_1,y_1),\cdots,(I_n,y_n),I_{n+1},\cdots,I_{n+m}\}$, where $I_i\in \mathbb{R}^{H\times W\times 3}$ refers to the $i$-th image, and its label $y_i=\{(b_{ik},c_{ik})\}_{k=1}^K$, where $K$  is the number of targets. Each target is represented by its bounding box coordinates $b_{ik}\in \mathbb{R}^4$ and the category label $c_{ik}\in \mathbb{R}^{\mathcal{C}_{\text{base}}}$. Our objective is to train a detector capable of detecting both base and novel categories, i.e., $\mathcal{C}_{\text{test}}=\mathcal{C}_{\text{base}}\cup\mathcal{C}_{\text{novel}}$, where $\mathcal{C}_{\text{base}}\cap\mathcal{C}_{\text{novel}}=\emptyset$.

\subsection{Open Vocabulary Object Detector}
\label{subsec:ovd}
\begin{figure}[t]
  \centering
   \includegraphics[width=0.93\textwidth]{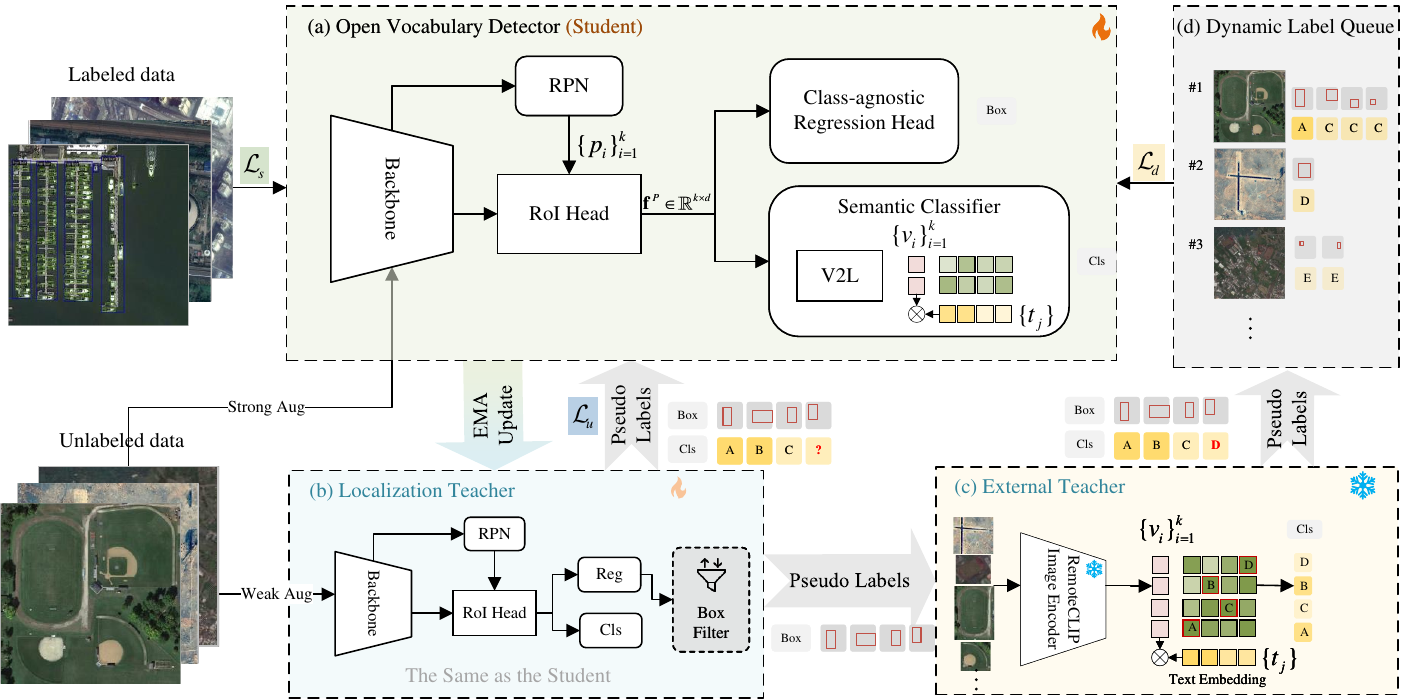}
   \caption{Overall architecture of CastDet. In each training iteration, the data batch consists of three data flow: labeled data with annotations, unlabeled data, and data sampled from the dynamic label queue. The labeled images are directly used for the student network training (\colorbox[rgb]{0.84,0.894,0.805}{$\mathcal{L}_s$}), while two sets of pseudo-labels of unlabeled data are predicted through the localization teacher and external teacher. One supervises the student (\colorbox[rgb]{0.726,0.816,0.898}{$\mathcal{L}_u$}), and the other is pushed into the dynamic label queue. Simultaneously, samples are randomly selected from the dynamic label queue to enhance the student's ability to detect novel targets (\colorbox[rgb]{0.99,0.94,0.805}{$\mathcal{L}_d$}).}
   \label{fig:framework}
\end{figure}
\textbf{Architecture Overview.} Fig. \ref{fig:framework} illustrates an overview of our CastDet framework. There is a student model and two teacher models: a localization teacher and an external teacher. The student model is the object detection model based on the Faster R-CNN architecture, with a modified class-agnostic bounding box regression head and a semantic classifier. The student model is trained on both the labeled samples and unlabeled samples with pseudo classification and bounding-box regression labels produced by the localization teacher and the dynamic label queue. The localization teacher is an exponential moving average (EMA) of the student model~\cite{tarvainen2017mean} so that it aggregates the history information during the training iterations to obtain better and stable representations, ensuring the quality of the pseudo-labels. During the training process, the localization teacher simultaneously generates two sets of pseudo-labels for the unlabeled images, one for the student model training and the other with pseudo boxes input to the external teacher for the pseudo-label generation. The external teacher is a frozen RemoteCLIP fundamental model~\cite{liu2023remoteclip} which is a vision-language model pre-trained on large-scale remote sensing image-text pairs following the CLIP framework, bearing strong open-vocabulary classification ability by comparing their image embedding and category embedding.  Furthermore, we employ a dynamic queue to store the pseudo-labels generated by the external teacher to facilitate maintaining high-quality pseudo-labels and balanced data sampling for the student model training.

\noindent
\textbf{Class-agnostic box regression head} is the bounding box regression branch, sharing the parameters for all categories, i.e., the prediction of the regression box $b_i\in \mathbb{R}^4$ instead of $b_i\in\mathbb{R}^{4|\mathcal{C}_{\text{test}}|}$, for each box $i$. As described in \cite{feng2022promptdet,zhou2022detic}, this approach can simplify the model and make it more versatile, allowing it to handle cases where the number of object categories is not fixed.

\noindent
\textbf{Semantic classifier head} aims to classify RoI regions (Region of Interest) beyond a predefined set of categories. We follow Detic\cite{zhou2022detic} to use the semantic embeddings for the category vocabulary as the weight of the last fully connected layer. By doing so, the prediction categories can be easily expanded. The semantic embeddings are generated by two steps: (1)  Filling the concept with a pre-defined prompt template ``\texttt{a photo of [\underline{category}]}''. (2) Encoding the text descriptions into semantic embeddings $t_j$ through the pre-trained text encoder of RemoteCLIP. Given a set of RoI features $\{v_i\}_{i=1}^k$, the 
prediction score is calculated as
\begin{equation}
\small
\label{equ:s_ij}
    \hat{s}_{ij}=\frac{v_i^T\cdot t_j}{\tau\left \| v_i \right \|\cdot \left \| t_j \right \|},
\end{equation}
where $\tau$ is the temperature parameter that controls the range of the logits in the softmax which is directly optimized during training as a log parameterized multiplicative scalar as in \cite{CLIP_radford2021learning}.

\subsection{Localization Teacher}
\label{sec:localization_teacher}
\noindent
\textbf{Exponential Moving Average.}
As we discussed before, the recall of region proposals for novel categories in aerial images is significantly lower than that in natural images (Fig.~\ref{fig:dataset_scale}). To tackle this problem, we employ a robust teacher for object discovery. In order to achieve open-vocabulary detection, the teacher needs to continuously update to learn how to discover and localize all possible novel categories. Thus, we adopt an interactive learning mechanism between the student and teacher model instead of a frozen teacher. Inspired from \cite{tarvainen2017mean, ge2020mutual, he2020momentum}, the teacher model is updated by an exponential moving average of the student model during training iterations.  The weights of the teacher $\theta^\prime$ are  updated as a weighted average of successive weights of the student $\theta$ at training iteration $t$:
\begin{equation}
\small
    \theta_t^\prime = \alpha\theta_{t-1}^\prime + (1-\alpha)\theta_t
\end{equation}
where $\alpha\in [0,1)$ is a momentum coefficient.

This brings three practical advantages over a frozen teacher: Firstly, the teacher can fully exploit the unlabeled data to improve the accuracy of the model with fewer annotation data; Secondly, it can aggregate the history information of the student model, thereby obtaining more robust predictions~\cite{tarvainen2017mean}; Thirdly, the approach achieves on-line learning and can scale to more novel concepts.

\noindent
\textbf{Consistency Training with Entropy Minimization.}
Given an unlabeled image, weak and strong augmentation are applied to it, serving as inputs for the localization teacher and student, respectively. We apply consistency training~\cite{berthelot2019mixmatch} that encourages the student to predict the same categories as the teacher for an unlabeled augmented input. Then, we minimize the cross entropy between these two predictions~\cite{sohn2020fixmatch}, i.e., $\min_\theta H(p_m(y|\theta),p_m(y|\theta^\prime))$. The training objective will be further discussed in Sec.~\ref{sec:hybrid_training}.

\begin{figure}[t]
  \centering
\includegraphics[width=\linewidth]{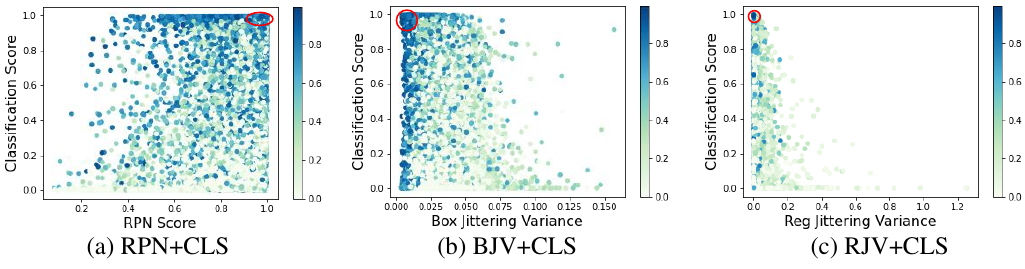}
\caption{Visualization of three types of box selection strategies. The figures shows the correlation among IoU, classification score, and (a) RPN score, (b) box-jittering variance, and (c) regression-jittering variance, respectively. Among them, IoU is represented by the color bar.}
   \label{fig:score_analysis}
\end{figure}

\noindent
\textbf{Box Selection Strategies.}
The primary task of the localization teacher is to determine the bounding box of the objects, of which the accuracy greatly impacts the quality of pseudo-labels generated by the external teacher. At this stage, we place a higher emphasis on the precision of pseudo-boxes, as student-teacher learning inherently yields favorable results with fewer labels~\cite{tarvainen2017mean}. Specifically, the localization teacher first generates a set of box candidates, which are then filtered based on their confidence scores before being presented to the external teacher. We compare several indicators, as shown in Fig.~\ref{fig:score_analysis}:
\begin{enumerate}
    \item RPN Score. This strategy filters out boxes with low RPN foreground confidence, a common approach adopted by most OVD methods~\cite{zhao2022exploiting,chen2023ovarnet}. 
    \item Box Jittering Variance (BJV). Box jittering means randomly sampling a set of jitter boxes around $b_i$ and predicting their refined box $\{\hat{b_{i,j}} \}$. The BJV is defined as $\bar{\sigma_i} = \frac{1}{4}\sum_{k=1}^4 \frac{\sigma_{ik}}{0.5(h_i+w_i)}$, where $\{\sigma_{ik}\}_{k=1}^4$, $h_i$, $w_i$ denote the standard derivation, height and width of the $i$-th boxes set, respectively~\cite{softteacher_xu2021end}.
    \item Regression Jittering Variance (RJV). Regression jittering means we iteratively put the predicted box into the regression branch for a more precise prediction. The RJV is defined as $\bar{\sigma_i} = \frac{1}{4}\sum_{k=1}^4 \frac{\sigma_{ik}^2}{(h_{-1}^2+w_{-1}^2)}$, where $\{\sigma_{ik}\}_{k=1}^4$ is the standard derivation of the $i$-th set of regression boxes, $h_{-1}$ and $w_{-1}$ are the height and width of the last regression box, respectively.
\end{enumerate}

\subsection{Dynamic Pseudo Label Queue}
\label{sec:dynamic_queue}
\begin{figure}[t]
  \centering
   \includegraphics[width=0.9\textwidth]{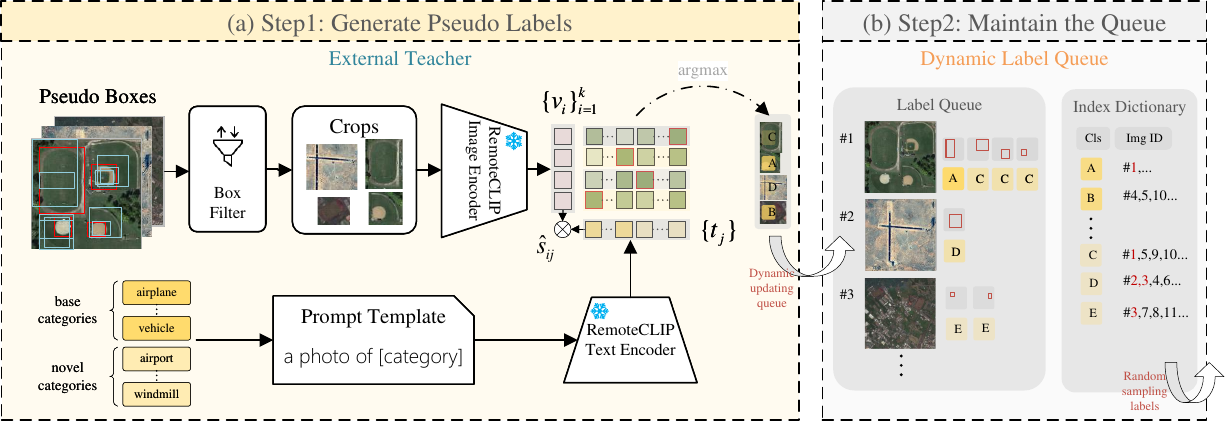}
   \caption{Workflow of dynamic label queue. \textbf{Step1:} filter certain high-quality proposal boxes generated by the localization teacher, and employ RemoteCLIP to classify corresponding crop images as pseudo labels. \textbf{Step2:} dynamically update those pseudo-labels into the queue, and randomly sample a batch of pseudo labels for the student training.}
   \label{fig:dynamic_queue}
\end{figure}
The workflow of the dynamic queue comprises two steps: 1) generating high-quality pseudo-labels through the external teacher and 2) dynamically updating the pseudo label queue and transferring data, as illustrated in Fig.~\ref{fig:dynamic_queue}.

\noindent
\textbf{Generate Pseudo Labels.}
Given an unlabeled image $I$ as input to the localization teacher, the RPN generates a set of proposals $\{p_i\}$. Subsequently, we extract the regional features $\mathbf{f}^P\in \mathbb{R}^{k\times d}$ of these proposals through RoI pooling and predict their coordinates $B=\{\hat{b}_i\}^{k}_{i=1}$ through the class-agnostic regression branch. However, directly inputting these boxes to RemoteCLIP for category prediction is computationally wasteful and redundant. Therefore, we use a box filter to select $k$ candidates, as discussed in Sec.~\ref{sec:localization_teacher}. Finally, we obtain a set of image crops $I^\prime=\{x^I_i\}_{i=1}^{k}$ by cropping the corresponding areas from the image.

For an image crop $x^I_i$, we first extract its visual feature $v_i$ via the visual encoder of RemoteCLIP. The text embeddings $t_j$ are generated as described in Sec.~\ref{subsec:ovd}. The prediction probability is performed by computing the softmax value for similarity between the visual and text embeddings:
\begin{equation}
\small
\label{equ:p_ij}
    \hat{p}_{ij}=\frac{e^{\hat{s}_{ij}}}{\sum_k e^{\hat{s}_{ik}}},
\end{equation}
where $\hat{s}_{ij}$ is calculated by Equ (\ref{equ:s_ij}).

To ensure the reliability of pseudo-labels, we filter the prediction probability $\hat{p}_{ij}$ with a relatively high threshold $p_0$, and push the image with positive samples $(I,\hat{y})$ into the dynamic label queue, where $\hat{y}=\{(\hat{b}_i,\hat{c}_i)\}_{i=1}^{\hat{k}}$ and $\hat{c}_i=\arg\max_j \hat{p}_{ij}$ denotes the prediction label.

\noindent
\textbf{Maintain the Queue.}
The dynamic label queue comprises a pseudo-label queue for storing image metadata (e.g., image path, labels, boxes, etc.), and an index dictionary to manage the mapping relationship between categories and image indexes, as depicted in Fig.~\ref{fig:dynamic_queue}(b). The label queue and index dictionary are dynamically updated through a continuous influx of pseudo-boxes generated by the localization teacher. 
Specifically, the same image is overwritten, and the images identified as non-existing objects previously are pushed into the queue during subsequent predictions. At the same time, the index dictionary is updated as \texttt{\{cls\_id:list[image\_ids]\}}.
This dynamic process enables the queue to accumulate richer and more accurate pseudo-labels as the model iterates.

Data transmission from the dynamic label queue to the student model is regulated by the index dictionary. Initially, the teachers iterate through all the unlabeled data and push the pseudo labels into the queue. Subsequently, images are randomly sampled from the \texttt{[image\_ids]} list of each category with a specified probability. The chosen images, along with their pseudo labels, are utilized for training the student model. This approach serves as an incentive to introduce knowledge about novel categories, fostering a positive feedback loop for both the student and localization teacher. Consequently, the entire system is driven to learn and discover novel targets.

\subsection{Hybrid Training}
\label{sec:hybrid_training}
The training process is depicted in Fig.~\ref{fig:framework}, with the overall loss comprising three components:
\begin{equation}
\small
    \mathcal{L} = \mathcal{L}_s+ \alpha\mathcal{L}_u + \beta\mathcal{L}_d,
\end{equation}
where $\mathcal{L}_s$, $\mathcal{L}_u$ and $\mathcal{L}_d$ denote supervised loss of labeled images, unsupervised loss of unlabeled images with pseudo boxes annotated by the localization teacher, and unsupervised loss of images sampled from the dynamic label queue, respectively.

\noindent
\textbf{Labeled Data Flow.} Given a batch of labeled data $\{(I_k,\{(b_i,c_i)\})\}$, we utilize the open-vocabulary detector to predict their coordinates $\{\hat{b}_i\}$ and prediction scores $\{\hat{s}_i\}$, as illustrated in Fig.~\ref{fig:framework}(a). The supervised loss is calculated as
\begin{equation}
\small
        \mathcal{L}_s = \frac{1}{N_b}\sum_{i=1}^{N_b}  \mathcal{L}_{\mathrm{cls}}(\hat{s}_i,c_i)+\frac{1}{N_{b}^{\mathrm{fg}}}\sum_{i=1}^{N_{b}^{\mathrm{fg}}} \mathcal{L}_{\mathrm{reg}}(\hat{b}_i, b_i)
\end{equation}
where $\mathcal{L}_{\mathrm{cls}}$ is the classification loss, $\mathcal{L}_{\mathrm{reg}}$ is the box regression loss, $N_b$ and $N_{b}^{\mathrm{fg}}$ denote the total number of proposals and the number of foreground proposals.

\noindent
\textbf{Unlabeled Data Flow.}
The unsupervised loss $\mathcal{L}_u$ is consists of two parts: classification loss $\mathcal{L}_u^{\mathrm{cls}}$ and box regression loss $\mathcal{L}_u^{\mathrm{reg}}$.
At the initial stages of the training process, the system struggles to detect novel categories. Directly filtering predictions by the RPN score or classification score would result in a large number of false negatives. Therefore, we assign a weight $w_j$ for the negative samples, which is the normalized contribution of the background prediction score of the $j$-th candidate. The classification loss is defined as:
\begin{equation}
\small
    \mathcal{L}_{u}^{\mathrm{cls}}=\frac{1}{N_{b}^{\mathrm{fg}}} \sum_{i=1}^{N_{b}^{\mathrm{fg}}} \mathcal{L}_{\mathrm{cls}}\left(\hat{s}_{i}, \hat{c}_i\right)+\sum_{j=1}^{N_{b}^{\mathrm{bg}}} w_{j} \mathcal{L}_{\mathrm{cls}}\left(\hat{s}_{j}, \hat{c}_j\right),
\end{equation}
where $N_{b}^{\mathrm{bg}}$ denotes the total number of background targets.

We apply the box selection strategy (in Sec.~\ref{sec:localization_teacher}) to filter candidates for training the regression branch. The regression loss is defined as:
\begin{equation}
\small
    \mathcal{L}_{u}^{\mathrm{reg}}=\frac{1}{N_{b}^{\mathrm{fg}}} \sum_{i=1}^{N_{b}^{\mathrm{fg}}} \mathcal{L}_{\mathrm{reg}}\left(\hat{b}_{i}^{\mathrm{fg}}, \hat{b}_i\right)
\end{equation}
where $\hat{b}_{i}^{\mathrm{fg}}$ and $\hat{b}_i$ denote the predicted foreground box and the assigned pseudo box, respectively.

\noindent
\textbf{Queue Data Flow.} In order to motivate the student-multi-teacher model to uncover novel objectives within the self-learning process, a specific number of images are randomly sampled from the dynamic label queue, as elaborated in Sec.~\ref{sec:dynamic_queue}. 
Given that the localization teacher is already responsible for instructing the student in the discovery and localization of targets, we exclusively compute the classification loss for these sampled images, aiming to infuse novel knowledge into the training of the student. The objective is formatted as follows:
\begin{equation}
\small
        \mathcal{L}_d = \frac{1}{N_b}\sum_{i=1}^{N_b} \mathcal{L}_{\mathrm{cls}}(\hat{s}_i,\hat{c}_i),
\end{equation}

\section{Experiments}
\subsection{Datasets and Settings}
\textbf{Datasets.} Due to the absence of dataset configurations specifically designed for OVD in aerial imagery currently, we follow the setup for generalized zero-shot-detection (GZSD) setting in VisDrone2023 Challenge\cite{VisDrone2023} to split the base and novel categories. We evaluate our CastDet on typical aerial datasets, including VisDroneZSD\cite{visdrone_zhu2021detection}, DOTA\cite{DOTA8578516}, NWPU VHR-10\cite{nwpu_cheng2016learning}. We utilize a subset of DIOR\cite{dior_li2020object} as supplemental unlabeled training data. For some comparison open-vocabulary detectors that require additional classification or caption data, we provide NWPU-RESISC45~\cite{NWPU_RESISC45_7891544} dataset as weak supervision. More details for datasets setting can be found in supplementary material.


\noindent
\textbf{Evaluation Metrics.}
We take the mean Average Precision (mAP), mean Average Recall (mAR) and Harmonic Mean (HM) metrics as our evaluation metrics. The mAP and mAR are averaged over an Intersection Over Union (IoU) value threshold of 0.5. Following \cite{VisDrone2023}, we utilize the Harmonic Mean (HM) as another metric to provide a comprehensive evaluation, which is defined as the overall mAP performance of base and novel categories, i.e.,
\begin{equation}
\small
    \mathrm{HM} = 2\frac{{\mathrm{mAP_{base}}}\cdot{\mathrm{mAP_{novel}}}}{{\mathrm{mAP_{base}}}+{\mathrm{mAP_{novel}}}} 
\end{equation}

\noindent
\textbf{Implementation Details.}
We implement our method with MMDetection toolbox\cite{mmdetection}. We employ Faster R-CNN\cite{ren2015faster} with ResNet50-C4~\cite{he2016deep} backbone as our detection framework. The model is initialized by a pre-trained Soft Teacher~\cite{softteacher_xu2021end} and R50-RemoteCLIP~\cite{liu2023remoteclip}, followed by 10k iterations training with a batch size of 12 on a single A6000 GPU. Stochastic Gradient Descent (SGD) is adopted as the optimizer with a learning rate of 0.01, and the momentum and weight decay parameters are configured to 0.9 and 0.0001, respectively. To maintain high-confidence pseudo-labels, we set the RPN foreground threshold and prediction probability threshold $p_0$ relatively high at 0.95 and 0.8, respectively.

\begin{table}[!htbp]
\caption{Effectiveness of hybrid training in enhancing novel category discovery. A comparative analysis of region proposal recall for different training strategies. S, LT, ET denote the student, localization teacher and external teacher, respectively.} 
\centering
\resizebox{0.75\linewidth}{!}{\begin{tabular*}{0.88\linewidth}{ccc|cccc|cccc}
\toprule
                   & &      &                                           &                                          &                                         &                              & \multicolumn{4}{c}{novel categories}                                                                                      \\
\multirow{-2}{*}{S} & \multirow{-2}{*}{LT} & \multirow{-2}{*}{ET} & \multirow{-2}{*}{mAR} & \multirow{-2}{*}{mAR$_{\mathrm{base}}$} & \multirow{-2}{*}{mAR$_{\mathrm{novel}}$} & \multirow{-2}{*}{HM}         & airport                      & bball.ct              & track.fld             & windmill                     \\ \hline
\Checkmark & &             & 41.7                                     & 46.0                                    & 24.7                                   & 32.2                & 22.1          & 29.7            & 40.6             & 6.5           \\
\Checkmark & \Checkmark &            & 60.1                                     & \textbf{63.2}                           & 47.7                                   & 54.4                & 31.8          & 80.2            & \textbf{73.0}    & 5.9           \\
\Checkmark & \Checkmark & \Checkmark          & \textbf{63.6}                            & 62.2                                    & \textbf{69.1}                          & \textbf{65.5}       & \textbf{72.1} & \textbf{91.8}   & 71.3             & \textbf{41.3}\\ 
 \bottomrule
\end{tabular*}}
  \label{tab:recall_bybrid}
\end{table}

\subsection{Ablation Study}
We conduct ablation studies on VisDroneZSD to validate the effectiveness of the proposed method. These studies include different training strategies, the dynamic label queue, box selection strategies, and label fraction experiments.

\noindent
\textbf{Effectiveness of Hybrid Training.}
To demonstrate that hybrid training can effectively guide the model to discover novel categories, we compare the recall of RPN the detector under three different training mechanisms: supervised pipeline (S), closed-vocabulary student-teacher semi-supervised pipeline (S+LT) and our open-vocabulary hybrid training (S+LT+ET).
For a fair comparison, our statistical recall is class-agnostic, i.e., as long as a target is proposed by RPN, it is considered to be successfully detected. As shown in Table \ref{tab:recall_bybrid}, the RPN recall significantly improves by our hybrid training for novel categories compared to the semi-supervised pipeline (e.g., 69.1 vs. 47.7 mAR$_\mathrm{novel}$). Supervised pipeline is poor at localizing novel categories, semi-supervised pipeline performs better on only some of the novel categories, while our hybrid training approach effectively improves the detection of novel categories.

\noindent
\textbf{Box Selection Strategy.} We compare the effects of different box selection strategies. The results are shown in Table~\ref{tab:ab7_box_selection}. The regression jittering strategy improves the performance by 4.5\% mAP$_\mathrm{novel}$ compared to the RPN Score strategy. As depicted in Fig.~\ref{fig:score_analysis}, it exhibits a stronger correlation with both the classification score and the IoU score. Therefore, it is possible to select more precise pseudo labels, which will have a beneficial impact on the training process.

\noindent
\textbf{Effectiveness of Dynamic Label Queue.}
To illustrate that the dynamic label queue can obtain high-quality pseudo-labels, we conduct experiments on whether to adopt a dynamic update strategy or whether to use a label queue for storing pseudo-labels.
In Table \ref{tab:ab2_dynamic_queue}, we can see substantial improvement achieved through applying dynamic updating or label queue techniques.

\noindent
\textbf{Label Fraction Experiments.} To further validate that our approach is suitable for the aerial scenarios with a limited amount of labeled data, we trained our model using 34\%, 50\%, and 100\% of the labelled data. The results are outlined in Table~\ref{tab:ab5_label_fraction}. Notably, even with a reduction to 34\% of the original labeled data, the performance did not drop significantly (e.g., 39.5\% mAP vs. 38.6\% mAP for 100\% and 34\% labeled data, respectively).

\begin{table*}[t]
\centering
\caption{Ablation experiments results. (a) Comparison on box selection strategies. (b) Evaluation on other remote sensing datasets. (c) Ablation study on fraction of labeled data.  (d) Evaluation on CLIPs (e.g., RemoteCLIP, CLIP) with ground-truth (GT) or RPN pseudo boxes. (e) Ablation study on dynamic label queue.}
\label{tab:ab_exps} 
\subfloat[
Box selection strategies.
\label{tab:ab7_box_selection}
]{
\centering
\begin{minipage}{0.30\linewidth}{\begin{center}
\resizebox{0.98\linewidth}{!}{
\begin{tabular}{c|cccc}
\toprule
Strategy      & mAP & mAP$_{\mathrm{base}}$ & mAP$_{\mathrm{novel}}$ & HM            \\ \hline
RPN Score     & 39.5          & 38.6            & 43.3             & 40.8          \\
Box Jittering & 39.2          & 38.0            & 43.6             & 40.6          \\
Reg Jittering & \textbf{40.7} & \textbf{39.0}   & \textbf{47.8}    & \textbf{42.9} \\ \bottomrule
\end{tabular}}
\end{center}}\end{minipage}
 }
\subfloat[Other Datasets.\label{tab:other_dataset_exp}
]{
\begin{minipage}{0.36\linewidth}{\begin{center}
\resizebox{0.98\linewidth}{!}{
\begin{tabular}{cc|cccc}
\toprule
Dataset     & \#External & mAP & mAP$_{\mathrm{base}}$ & mAP$_{\mathrm{novel}}$ & HM   \\ \hline
VisDroneZSD & DIOR       & 40.5 & 39.0                   & 46.3                    & 42.3 \\
NWPU VHR-10       & DIOR       & 87.6 & 86.8                   & 90.6                    & 88.7 \\
DOTA        & DIOR       & 58.6 & 61.4                   & 40.2                    & 48.6
\\ \bottomrule
\end{tabular}
}
\end{center}}\end{minipage}
}
\subfloat[
Label fraction.
\label{tab:ab5_label_fraction}
]{
\begin{minipage}{0.30\linewidth}{\begin{center}
\resizebox{0.98\linewidth}{!}{\begin{tabular}{c|cccc}
\toprule
Label Fraction & mAP & mAP$_{\mathrm{base}}$ & mAP$_{\mathrm{novel}}$ & HM             \\ \hline
34\%           & 38.6          & 38.0            & 41.0             & 39.5          \\
50\%           & 38.8          & 37.7            & \textbf{43.4}    & 40.4          \\
100\%          & \textbf{39.5} & \textbf{38.6}   & 43.3             & \textbf{40.8} \\ \bottomrule
\end{tabular}
}
\end{center}}\end{minipage}
}
\\
\centering
\subfloat[
Comparison with CLIPs.
\label{tab:com2clip}
]{
\centering
\begin{minipage}{0.6\linewidth}{\begin{center}
\resizebox{0.98\linewidth}{!}{
\begin{tabular}{cc|cccc|cccc}
\toprule
       &          & \multicolumn{4}{c|}{VisDrone}                                                  & \multicolumn{4}{c}{COCO}                                                       \\ \hline
Method & Proposal & mAP           & mAP$_{\mathrm{base}}$ & mAP$_{\mathrm{novel}}$ & HM            & mAP           & mAP$_{\mathrm{base}}$ & mAP$_{\mathrm{novel}}$ & HM            \\ \hline
CLIP   & GT       & 37.6          & 30.4                  & \textbf{66.4}          & 41.7          & 39.8          & 39.1                  & \textbf{41.5}          & 40.3          \\
Ours   & GT       & \textbf{47.3} & \textbf{46.6}         & 49.9                   & \textbf{48.2} & \textbf{51.8} & \textbf{55.5}         & 41.4                   & \textbf{47.4} \\ \hline
CLIP   & RPN      & 11.6          & 9.7                   & 19.3                   & 12.9          & 14.7          & 15.2                  & 13.2                   & 14.2          \\
Ours   & RPN      & \textbf{38.1} & \textbf{36.6}         & \textbf{44.2}          & \textbf{40.0} & \textbf{37.0} & \textbf{40.6}         & \textbf{27.1}          & \textbf{32.5} \\ \bottomrule
\end{tabular}
}
\end{center}}\end{minipage}
}
\subfloat[
Dynamic Label Queue.
\label{tab:ab2_dynamic_queue}
]{
\begin{minipage}{0.38\linewidth}{\begin{center}
\resizebox{0.99\linewidth}{!}{
\begin{tabular}{cc|cccc}
\toprule
Dynamic                   & Queue                     & mAP & mAP$_{\mathrm{base}}$ & mAP$_{\mathrm{novel}}$ & HM   \\ \hline
                          &                           & 11.6  & 9.7                     & 19.3                     & 12.9 \\
\Checkmark &                           & 37.7 & 36.3 & \textbf{43.5} & 39.6 \\
                          & \Checkmark & 38.1 & 37.2 & 41.5 & 39.2 \\
\Checkmark & \Checkmark & \textbf{39.5} & \textbf{38.6} & 43.3 & \textbf{40.8} 
\\ \bottomrule
\end{tabular}
}
\end{center}}\end{minipage}
}

\end{table*}

\subsection{Comparison with the RemoteCLIP}
We conducted comparison experiments with RemoteCLIP~\cite{liu2023remoteclip} and CLIP~\cite{CLIP_radford2021learning}.
The corresponding results are presented in Table \ref{tab:com2clip}, revealing that CastDet exhibits an enhanced open-vocabulary classification capability compared to RemoteCLIP, i.e., from 37.6\% to 47.3\% mAP.  Moreover, CastDet demonstrates a more significant performance improvement when validating RPN's proposals, e.g., from 11.6\% to 38.1\% mAP. This indicates that our training approach effectively enhances novel object discovery and classification abilities. We further conduct experiments on COCO~\cite{lin2014microsoft}, and observe significant improvements with our approach on natural images as well.

\subsection{Comparison with the State-of-the-Art}
In OVD, auxiliary weak supervision may be used, such as: \textbf{1)} using both base and novel categories ($\mathcal{T}_{\mathrm{novel}}$)  with unlabeled images for knowledge distillation or pseudo-labeling, such as ViLD~\cite{gu2021open} and OV-DETR~\cite{zang2022open}, aligning with our setting; and \textbf{2)} using classification or caption data  ($\mathcal{T}_{\mathrm{cls}}$) that may contain novel concepts, such as Detic~\cite{zhou2022detic}, etc~\cite{Cheng2024YOLOWorld,liu2023groundingdino,9879567glip,wu2023baron}, therefore we provide an additional classification dataset NWPU-RESISC45~\cite{NWPU_RESISC45_7891544} for these methods. We compare the proposed method with above SOTA open-vocabulary detectors on VisDroneZSD dataset, as shown in Table \ref{tab:com2best}. Our method outperforms the previous SOTA by 21.0\% mAP$_{\mathrm{novel}}$. To further evaluate the open-vocabulary generalization capability of our method, we construct a large concept pool for self-training without requiring novel classes to be pre-known. The results in Table \ref{tab:com2best} show comparable performance. The visualization results can be found in supplementary material.

\begin{table*}[t]
\caption{Comparison with the state-of-the-art detectors.  $\mathcal{T}_{\mathrm{novel}}$ indicates whether the novel classes need to be pre-known.  $\mathcal{T}_{\mathrm{cls}}$ denotes whether extra classification or caption datasets are required during training.  $\dagger$: The best result on the zero-shot object detection leaderboard of VisDrone2023. $\ddagger$: The results of our own implementation, we replace the CLIP of ViLD with RemoteCLIP.}
\centering
\resizebox{1.\linewidth}{!}{
\begin{tabular*}{1.35\linewidth}{c|cccccc|cccc}
\toprule
Method        & Backbone     & External    & Labeled Data             & Task & $\mathcal{T}_{\mathrm{novel}}$  & $\mathcal{T}_{\mathrm{cls}}$ & mAP & mAP$_{\mathrm{base}}$ & mAP$_{\mathrm{novel}}$ & HM   \\ \hline
MultiModel    & -            & -             & VisDroneZSD              & ZSD  & \xmark & \xmark & -    & -                      & -                       & 26.7$^\dagger$ \\
ViLD$^\ddagger$~\cite{gu2021open}         & RN50 & DIOR      & VisDroneZSD              & OVD  & \cmark & \xmark    & 25.6          & 28.5            & 14.2             & 19.0 \\
OV-DETR~\cite{zang2022open} & RN50 & DIOR & VisDroneZSD & OVD & \cmark & \xmark & 25.6 & 25.6 & 25.5 & 25.6 \\
Detic~\cite{zhou2022detic}         & RN50    & NWPU-RESISC45 & VisDroneZSD              & OVD  & \xmark & \cmark & 16.8          & 19.8            & 4.8              & 7.7 \\
GroundingDINO~\cite{liu2023groundingdino} & RN50 & NWPU-RESISC45 & VisDroneZSD & OVD  &\xmark &\cmark & 33.0 & \underline{40.5} & 3.3 & 6.1 \\
GLIP~\cite{9879567glip} & Swin-T & NWPU-RESISC45 & VisDroneZSD & OVD  &\xmark &\cmark & 33.8 & \textbf{41.0} & 5.4 & 9.5 \\
BARON~\cite{wu2023baron} & RN50 & NWPU-RESISC45 & VisDroneZSD & OVD &\xmark &\cmark & 27.4 & 29.4 & 19.5 & 23.5 \\
YOLO-World~\cite{Cheng2024YOLOWorld} & YOLOv8-M & NWPU-RESISC45 & VisDroneZSD & OVD &\xmark &\cmark  &  32.9 & 39.1 & 8.5 & 13.9 \\
\hline
CastDet (\textit{Ours}) & RN50 & DIOR          &  VisDroneZSD & OVD  &\xmark &\xmark & \underline{38.5} & 36.5   & \textbf{46.5}    & \underline{40.9}
\\
CastDet (\textit{Ours}) & RN50 & DIOR          &  VisDroneZSD & OVD  &\cmark &\xmark & \textbf{40.5} & 39.0   & \underline{46.3}    & \textbf{42.3 }
\\ \bottomrule
\end{tabular*}
}
  \label{tab:com2best}
\end{table*}


\subsection{Evaluation on Other Dataset}
We further validate our approach on the NWPU VHR-10\cite{nwpu_cheng2016learning} and DOTA\cite{DOTA8578516}. Our method achieves 87.6\% mAP on NWPU VHR-10 and 58.6\% mAP on DOTA, as shown in Table \ref{tab:other_dataset_exp}.

\section{Conclusion}
We propose CastDet, a CLIP-activated student-teacher detector designed for open-vocabulary aerial detection. Specifically, we introduce a localization teacher with several box selection strategies (e.g., RPN, BJV, RJV) for novel category discovery. Then, we incorporate RemoteCLIP to acquire limited yet reliable knowledge, serving as an external incentive for student-teacher interactive self-learning. We further introduce a dynamic label queue to store and dynamically update information to obtain richer and more accurate external knowledge. Additionally, we propose a hybrid training approach to simultaneously train multiple data streams, facilitating collaborative training of various sub-modules. With these improvements, CastDet achieves a 40.5\% mAP on VisDroneZSD. To the best of our knowledge,  this marks the first work on OVD in aerial images. We aspire to lay a foundation for subsequent research in this domain.

\section*{Acknowledgements}
This work was supported by the National Natural Science Foundation of China (62071333), Fundamental Research Funds for the Central Universities (22120220654), and the Science and Technology on Near-Surface Detection Laboratory (6142414221606). 

%
%

\begin{thebibliography}{10}
\providecommand{\url}[1]{\texttt{#1}}
\providecommand{\urlprefix}{URL }
\providecommand{\doi}[1]{https://doi.org/#1}

\bibitem{berthelot2019mixmatch}
Berthelot, D., Carlini, N., Goodfellow, I., Papernot, N., Oliver, A., Raffel, C.A.: Mixmatch: A holistic approach to semi-supervised learning. Advances in neural information processing systems  \textbf{32} (2019)

\bibitem{carion2020end}
Carion, N., Massa, F., Synnaeve, G., Usunier, N., Kirillov, A., Zagoruyko, S.: End-to-end object detection with transformers. In: European conference on computer vision. pp. 213--229. Springer (2020)

\bibitem{mmdetection}
Chen, K., Wang, J., Pang, J., Cao, Y., Xiong, Y., Li, X., Sun, S., Feng, W., Liu, Z., Xu, J., Zhang, Z., Cheng, D., Zhu, C., Cheng, T., Zhao, Q., Li, B., Lu, X., Zhu, R., Wu, Y., Dai, J., Wang, J., Shi, J., Ouyang, W., Loy, C.C., Lin, D.: {MMDetection}: Open mmlab detection toolbox and benchmark. arXiv preprint arXiv:1906.07155  (2019)

\bibitem{chen2023ovarnet}
Chen, K., Jiang, X., Hu, Y., Tang, X., Gao, Y., Chen, J., Xie, W.: Ovarnet: Towards open-vocabulary object attribute recognition. In: Proceedings of the IEEE/CVF Conference on Computer Vision and Pattern Recognition. pp. 23518--23527 (2023)

\bibitem{NWPU_RESISC45_7891544}
Cheng, G., Han, J., Lu, X.: Remote sensing image scene classification: Benchmark and state of the art. Proceedings of the IEEE  \textbf{105}(10),  1865--1883 (2017). \doi{10.1109/JPROC.2017.2675998}

\bibitem{nwpu_cheng2016learning}
Cheng, G., Zhou, P., Han, J.: Learning rotation-invariant convolutional neural networks for object detection in vhr optical remote sensing images. IEEE Transactions on Geoscience and Remote Sensing  \textbf{54}(12),  7405--7415 (2016)

\bibitem{Cheng2024YOLOWorld}
Cheng, T., Song, L., Ge, Y., Liu, W., Wang, X., Shan, Y.: Yolo-world: Real-time open-vocabulary object detection. In: Proc. IEEE Conf. Computer Vision and Pattern Recognition (CVPR) (2024)

\bibitem{5206848imagenet}
Deng, J., Dong, W., Socher, R., Li, L.J., Li, K., Fei-Fei, L.: Imagenet: A large-scale hierarchical image database. In: 2009 IEEE Conference on Computer Vision and Pattern Recognition. pp. 248--255 (2009). \doi{10.1109/CVPR.2009.5206848}

\bibitem{ding2019learning}
Ding, J., Xue, N., Long, Y., Xia, G.S., Lu, Q.: Learning roi transformer for oriented object detection in aerial images. In: Proceedings of the IEEE/CVF Conference on Computer Vision and Pattern Recognition. pp. 2849--2858 (2019)

\bibitem{du2022learning}
Du, Y., Wei, F., Zhang, Z., Shi, M., Gao, Y., Li, G.: Learning to prompt for open-vocabulary object detection with vision-language model. In: Proceedings of the IEEE/CVF Conference on Computer Vision and Pattern Recognition. pp. 14084--14093 (2022)

\bibitem{feng2022promptdet}
Feng, C., Zhong, Y., Jie, Z., Chu, X., Ren, H., Wei, X., Xie, W., Ma, L.: Promptdet: Towards open-vocabulary detection using uncurated images. In: European Conference on Computer Vision. pp. 701--717. Springer (2022)

\bibitem{ge2020mutual}
Ge, Y., Chen, D., Li, H.: Mutual mean-teaching: Pseudo label refinery for unsupervised domain adaptation on person re-identification. arXiv preprint arXiv:2001.01526  (2020)

\bibitem{gu2021open}
Gu, X., Lin, T.Y., Kuo, W., Cui, Y.: Open-vocabulary object detection via vision and language knowledge distillation. arXiv preprint arXiv:2104.13921  (2021)

\bibitem{gupta2019lvis}
Gupta, A., Dollar, P., Girshick, R.: Lvis: A dataset for large vocabulary instance segmentation. In: Proceedings of the IEEE/CVF conference on computer vision and pattern recognition. pp. 5356--5364 (2019)

\bibitem{he2020momentum}
He, K., Fan, H., Wu, Y., Xie, S., Girshick, R.: Momentum contrast for unsupervised visual representation learning. In: Proceedings of the IEEE/CVF conference on computer vision and pattern recognition. pp. 9729--9738 (2020)

\bibitem{he2016deep}
He, K., Zhang, X., Ren, S., Sun, J.: Deep residual learning for image recognition. In: Proceedings of the IEEE conference on computer vision and pattern recognition. pp. 770--778 (2016)

\bibitem{Align_jia2021scaling}
Jia, C., Yang, Y., Xia, Y., Chen, Y.T., Parekh, Z., Pham, H., Le, Q., Sung, Y.H., Li, Z., Duerig, T.: Scaling up visual and vision-language representation learning with noisy text supervision. In: International conference on machine learning. pp. 4904--4916. PMLR (2021)

\bibitem{dior_li2020object}
Li, K., Wan, G., Cheng, G., Meng, L., Han, J.: Object detection in optical remote sensing images: A survey and a new benchmark. ISPRS journal of photogrammetry and remote sensing  \textbf{159},  296--307 (2020)

\bibitem{9879567glip}
Li, L.H., Zhang, P., Zhang, H., Yang, J., Li, C., Zhong, Y., Wang, L., Yuan, L., Zhang, L., Hwang, J.N., Chang, K.W., Gao, J.: Grounded language-image pre-training. In: 2022 IEEE/CVF Conference on Computer Vision and Pattern Recognition (CVPR). pp. 10955--10965 (2022). \doi{10.1109/CVPR52688.2022.01069}

\bibitem{retina_lin2017focal}
Lin, T.Y., Goyal, P., Girshick, R., He, K., Doll{\'a}r, P.: Focal loss for dense object detection. In: Proceedings of the IEEE international conference on computer vision. pp. 2980--2988 (2017)

\bibitem{lin2014microsoft}
Lin, T.Y., Maire, M., Belongie, S., Hays, J., Perona, P., Ramanan, D., Doll{\'a}r, P., Zitnick, C.L.: Microsoft coco: Common objects in context. In: Computer Vision--ECCV 2014: 13th European Conference, Zurich, Switzerland, September 6-12, 2014, Proceedings, Part V 13. pp. 740--755. Springer (2014)

\bibitem{liu2023remoteclip}
Liu, F., Chen, D., Guan, Z., Zhou, X., Zhu, J., Zhou, J.: Remoteclip: A vision language foundation model for remote sensing. arXiv preprint arXiv:2306.11029  (2023)

\bibitem{liu2023groundingdino}
Liu, S., Zeng, Z., Ren, T., Li, F., Zhang, H., Yang, J., Li, C., Yang, J., Su, H., Zhu, J., et~al.: Grounding dino: Marrying dino with grounded pre-training for open-set object detection. arXiv preprint arXiv:2303.05499  (2023)

\bibitem{VisDrone2023}
team at Lab~of Machine~Learning, A., Mining, D.: Zero-shot object detection challenge (2023), \url{http://aiskyeye.com/challenge-2023/zero-shot-object-detection/}, Last accessed on 2023-11-09

\bibitem{qian2022rsdet++}
Qian, W., Yang, X., Peng, S., Zhang, X., Yan, J.: Rsdet++: Point-based modulated loss for more accurate rotated object detection. IEEE Transactions on Circuits and Systems for Video Technology  \textbf{32}(11),  7869--7879 (2022)

\bibitem{CLIP_radford2021learning}
Radford, A., Kim, J.W., Hallacy, C., Ramesh, A., Goh, G., Agarwal, S., Sastry, G., Askell, A., Mishkin, P., Clark, J., et~al.: Learning transferable visual models from natural language supervision. In: International conference on machine learning. pp. 8748--8763. PMLR (2021)

\bibitem{yolo_redmon2016you}
Redmon, J., Divvala, S., Girshick, R., Farhadi, A.: You only look once: Unified, real-time object detection. In: Proceedings of the IEEE conference on computer vision and pattern recognition. pp. 779--788 (2016)

\bibitem{reilly2010detection}
Reilly, V., Idrees, H., Shah, M.: Detection and tracking of large number of targets in wide area surveillance. In: Computer Vision--ECCV 2010: 11th European Conference on Computer Vision, Heraklion, Crete, Greece, September 5-11, 2010, Proceedings, Part III 11. pp. 186--199. Springer (2010)

\bibitem{ren2015faster}
Ren, S., He, K., Girshick, R., Sun, J.: Faster r-cnn: Towards real-time object detection with region proposal networks. Advances in neural information processing systems  \textbf{28} (2015)

\bibitem{sadgrove2018real}
Sadgrove, E.J., Falzon, G., Miron, D., Lamb, D.W.: Real-time object detection in agricultural/remote environments using the multiple-expert colour feature extreme learning machine (mec-elm). Computers in Industry  \textbf{98},  183--191 (2018)

\bibitem{sohn2020fixmatch}
Sohn, K., Berthelot, D., Carlini, N., Zhang, Z., Zhang, H., Raffel, C.A., Cubuk, E.D., Kurakin, A., Li, C.L.: Fixmatch: Simplifying semi-supervised learning with consistency and confidence. Advances in neural information processing systems  \textbf{33},  596--608 (2020)

\bibitem{7926624}
Sommer, L.W., Schuchert, T., Beyerer, J.: Fast deep vehicle detection in aerial images. In: 2017 IEEE Winter Conference on Applications of Computer Vision (WACV). pp. 311--319 (2017). \doi{10.1109/WACV.2017.41}

\bibitem{tarvainen2017mean}
Tarvainen, A., Valpola, H.: Mean teachers are better role models: Weight-averaged consistency targets improve semi-supervised deep learning results. Advances in neural information processing systems  \textbf{30} (2017)

\bibitem{wang2021knowledge}
Wang, L., Yoon, K.J.: Knowledge distillation and student-teacher learning for visual intelligence: A review and new outlooks. IEEE transactions on pattern analysis and machine intelligence  \textbf{44}(6),  3048--3068 (2021)

\bibitem{wu2023baron}
Wu, S., Zhang, W., Jin, S., Liu, W., Loy, C.C.: Aligning bag of regions for open-vocabulary object detection. In: CVPR (2023)

\bibitem{wu2023cora}
Wu, X., Zhu, F., Zhao, R., Li, H.: Cora: Adapting clip for open-vocabulary detection with region prompting and anchor pre-matching. In: Proceedings of the IEEE/CVF Conference on Computer Vision and Pattern Recognition. pp. 7031--7040 (2023)

\bibitem{DOTA8578516}
Xia, G.S., Bai, X., Ding, J., Zhu, Z., Belongie, S., Luo, J., Datcu, M., Pelillo, M., Zhang, L.: Dota: A large-scale dataset for object detection in aerial images. In: 2018 IEEE/CVF Conference on Computer Vision and Pattern Recognition. pp. 3974--3983 (2018). \doi{10.1109/CVPR.2018.00418}

\bibitem{softteacher_xu2021end}
Xu, M., Zhang, Z., Hu, H., Wang, J., Wang, L., Wei, F., Bai, X., Liu, Z.: End-to-end semi-supervised object detection with soft teacher. In: Proceedings of the IEEE/CVF International Conference on Computer Vision. pp. 3060--3069 (2021)

\bibitem{yang2019clustered}
Yang, F., Fan, H., Chu, P., Blasch, E., Ling, H.: Clustered object detection in aerial images. In: Proceedings of the IEEE/CVF international conference on computer vision. pp. 8311--8320 (2019)

\bibitem{yang2021r3det}
Yang, X., Yan, J., Feng, Z., He, T.: R3det: Refined single-stage detector with feature refinement for rotating object. In: Proceedings of the AAAI conference on artificial intelligence. vol.~35, pp. 3163--3171 (2021)

\bibitem{yang2019scrdet}
Yang, X., Yang, J., Yan, J., Zhang, Y., Zhang, T., Guo, Z., Sun, X., Fu, K.: Scrdet: Towards more robust detection for small, cluttered and rotated objects. In: Proceedings of the IEEE/CVF international conference on computer vision. pp. 8232--8241 (2019)

\bibitem{yao2022detclip}
Yao, L., Han, J., Wen, Y., Liang, X., Xu, D., Zhang, W., Li, Z., Xu, C., Xu, H.: Detclip: Dictionary-enriched visual-concept paralleled pre-training for open-world detection. Advances in Neural Information Processing Systems  \textbf{35},  9125--9138 (2022)

\bibitem{zang2022open}
Zang, Y., Li, W., Zhou, K., Huang, C., Loy, C.C.: Open-vocabulary detr with conditional matching. In: European Conference on Computer Vision. pp. 106--122. Springer (2022)

\bibitem{zareian2021open}
Zareian, A., Rosa, K.D., Hu, D.H., Chang, S.F.: Open-vocabulary object detection using captions. In: Proceedings of the IEEE/CVF Conference on Computer Vision and Pattern Recognition. pp. 14393--14402 (2021)

\bibitem{Zhang2023}
Zhang, X., Zhang, T., Wang, G., Zhu, P., Tang, X., Jia, X., Jiao, L.: Remote sensing object detection meets deep learning: A metareview of challenges and advances. IEEE Geoscience and Remote Sensing Magazine  \textbf{11}(4),  8--44 (2023). \doi{10.1109/MGRS.2023.3312347}

\bibitem{zhao2022exploiting}
Zhao, S., Zhang, Z., Schulter, S., Zhao, L., Vijay~Kumar, B., Stathopoulos, A., Chandraker, M., Metaxas, D.N.: Exploiting unlabeled data with vision and language models for object detection. In: European Conference on Computer Vision. pp. 159--175. Springer (2022)

\bibitem{zhao2003car}
Zhao, T., Nevatia, R.: Car detection in low resolution aerial images. Image and vision computing  \textbf{21}(8),  693--703 (2003)

\bibitem{zhong2022regionclip}
Zhong, Y., Yang, J., Zhang, P., Li, C., Codella, N., Li, L.H., Zhou, L., Dai, X., Yuan, L., Li, Y., et~al.: Regionclip: Region-based language-image pretraining. In: Proceedings of the IEEE/CVF Conference on Computer Vision and Pattern Recognition. pp. 16793--16803 (2022)

\bibitem{zhou2022detic}
Zhou, X., Girdhar, R., Joulin, A., Kr{\"a}henb{\"u}hl, P., Misra, I.: Detecting twenty-thousand classes using image-level supervision. In: European Conference on Computer Vision. pp. 350--368. Springer (2022)

\bibitem{visdrone_zhu2021detection}
Zhu, P., Wen, L., Du, D., Bian, X., Fan, H., Hu, Q., Ling, H.: Detection and tracking meet drones challenge. IEEE Transactions on Pattern Analysis and Machine Intelligence  \textbf{44}(11),  7380--7399 (2021)

\bibitem{levir_zou2017random}
Zou, Z., Shi, Z.: Random access memories: A new paradigm for target detection in high resolution aerial remote sensing images. IEEE Transactions on Image Processing  \textbf{27}(3),  1100--1111 (2017)

\end{thebibliography}



\section*{Supplementary material}

# Supplemental Materials

## A  Implementation Details

### A.1  Open-vocabulary Aerial Detection Benchmark Details

Due to the lack of open-vocabulary aerial datasets benchmark setting, we follow the generalized zero shot detection (GZSD) setting of VisDrone2023 Challenge [11] to divide the classes into base ($\mathcal{C}_{base}$) and novel ($\mathcal{C}_{novel}$) classes. This gives us 16 base classes and 4 novel classes for VisDroneZSD [11], 13 base classes and 2 novel classes for DOTA [12], 8 base classes and 2 novel classes for NWPU VHR-10 [3], as detailed in Table 1∼2. We assess our method using horizontal bounding boxes of each dataset, and the images without novel classes are employed as labeled data for training.

**Table 1:** Base/Novel Split on Aerial Datasets.

| Dataset | Base Categories | Novel Categories |
|---|---|---|
| VisDrone ZSD [11] | airplane, baseballfield, bridge, chimney, dam, expressway Service area, expressway toll station, golffield, harbor, overpass, ship, stadium, storagetank, tenniscourt, trainstation, vehicle | airport, basketballcourt, groundtrackfield, windmill |
| DOTA [12] | plane, baseballfield, bridge, small-vehicle, large-vehicle, ship, tenniscourt, storagetank, soccerballfield, roundabout, harbor, swimmingpool, helicopter | groundtrackfield, basketballcourt |
| NWPU VHR-10 [3] | airplane, ship, storage tank, baseball diamond, tennis court, harbor, bridge, vehicle | ground track field, basketball court |

**Table 2:** A summary of datasets used in our experiments. †: DIOR serves as supplementary unlabeled training data, while NWPU-RESISC45 is employed as extra weak supervision for comparison experiments. ‡: We crop the original images of the DOTA dataset into $800 \times 800$ patches with an overlap of 100.

| | Dataset | Image width | # Images | | | | # Categories | | |
|---|---|---|---|---|---|---|---|---|---|
| | | | total | labeled | unlabeled | test | total | base | novel |
| Det | VisDroneZSD [11] | 800 | 13067 | 8730 | - | 3337 | 20 | 16 | 4 |
| | NWPU VHR-10 [3] | ∼1000 | 800 | 205 | - | 445 | 10 | 8 | 2 |
| | DOTA [12] | 800∼4000 | ‡2806→15194 | 10045 | - | 5149 | 15 | 13 | 2 |
| | †DIOR [8] | 800 | 23463 | - | 8726 | - | 20 | - | - |
| Cls | †NWPU-RESISC45 [2] | 256 | 31500 | - | 2100 | - | 45 | - | - |

## A.2 COCO Benchmark Details

**Dataset setting.** We further evaluate our method using natural images, e.g., COCO [9]. Following [5,6,14,15], we choose 48 base classes and 17 novel classes from the 80 COCO classes. The train set remains consistent with COCO2017. During the supervised training process, only images containing at least one base class are used as labeled data. The novel classes in those images are ignored. In the hybrid training procedure, the training data include labeled data and unlabeled data. The labeled data only includes base categories, while the unlabeled data consists of the remaining data in the training set, and the ImageNet [4] dataset.

**Training details.** Following [5], we utilize the Mask R-CNN R50-FPN [7] for pre-training a class-agnostic region proposal generator. The model is trained with 48 base categories, employing a batch size of 16 on 2 A6000 GPUs, for a 12-epoch training cycle (1x schedule). Then, we initialize the proposal generator (i.e., backbone and RPN) of CastDet's student model and localization teacher model with the pre-trained Mask R-CNN. The model undergoes self-training for $30k$ iterations on a single A6000 GPU, utilizing a combination of labeled and unlabeled data.

**ViLD baseline details.** The original ViLD [6] is trained from scratch for $180k$ iterations of batch size 256 on multiple TPU devices, employing input images of size $1024\times1024$ with large-scale jittering augmentation. To compare with ViLD, we follow the training process outlined in their paper and released code, conducting experiments under a similar experimental setup as ours. In line with [6], we employ our pre-trained Mask R-CNN R50-FPN (1x schedule, batch size $16\times2$) to generate 1000 proposals. Subsequently, we obtain CLIP classification scores for the cropped regions of these proposals, perform Non-Maximum Suppression (NMS) with a threshold of 0.6, and retain the top 300 predictions as pseudos, including bounding boxes and visual features. For training ViLD, we initialize ViLD with Mask R-CNN and employ the same SGD optimizer with a learning rate of 0.01 and a batch size of 16 for $30k$ iterations, consistent with our setting.

# B   Open-vocabulary COCO Results

We compare our method with other previous works on COCO benchmark in Table 3. Limited by computational resources, we pretrain our model for only 12 epochs with base classes and conduct $30k$ iterations for hybrid training on no more than 2 GPUs. Despite limited resources, our method achieves impressive results in novel category detection, attaining a 30.3% mAP$_{novel}$, showing its effectiveness in detecting open-vocabulary categories. This underscores the efficiency and strength of our approach in challenging scenarios with resource constraints.

**Table 3:** Results on open-vocabulary COCO. †: Results quoted from [14]. ‡: The results of our own implementation, under the same experimental setup as ours. ∗: Results quoted from the original paper.

| | mAP$_{novel}$ | mAP$_{base}$ | mAP50$_{all}$ | HM |
|---|---|---|---|---|
| Mask R-CNN [7] | - | 51.4 | - | - |
| WSDDN [1]† | 19.7 | 19.6 | 19.6 | 19.6 |
| Cap2Det [13]† | 20.3 | 20.1 | 20.1 | 20.2 |
| OVR-CNN [14]∗ | 22.8 | 46.0 | 39.9 | 30.5 |
| Detic [15]∗ | 24.1 | 52.0 | 44.7 | 32.9 |
| ViLD [6]‡ | 12.9 | 38.8 | 32.0 | 19.3 |
| PromptDet [5]∗ | 26.6 | 59.1 | 50.6 | 36.7 |
| CastDet (Ours) | **30.3** | 47.4 | 42.9 | 37.0 |

## C   Analysis on Dynamic Label Queue

To demonstrate that the dynamic label queue can maintain richer and more accurate pseudo-labels throughout the training process, we evaluated precision and recall of pseudo-labels w.r.t. ground truth at different iterations on VisDroneZSD dataset. We set up a warm-up stage for $2k$ iterations, after which the pseudos in the label queue can be dynamically updated. As illustrated in Fig. 1, with the model iterations, the precision-recall (PR) curve expands outward (Fig. 1 (a)), and both mAP50$_{novel}$ and mAR@10$_{novel}$ of pseudo-labels show significant improvements compared to the initial state. However, with a static queue, the values of mAP50$_{novel}$ and mAR@10$_{novel}$ are likely to remain low, i.e., the same as in iter $2k$ (Fig. 1 (b)).

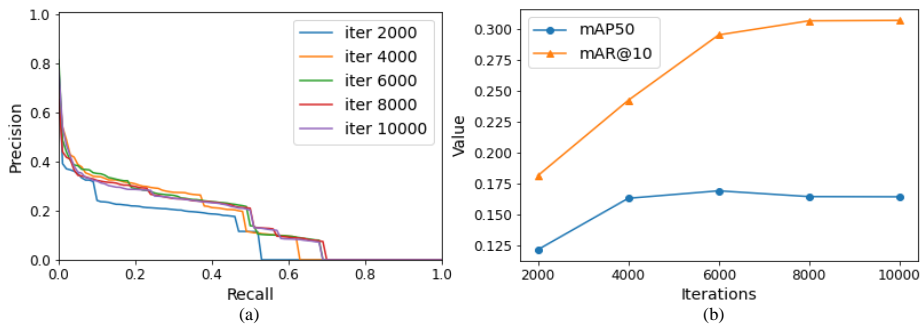

**Fig. 1:** Precision and recall statistics for novel categories of pseudo-labels in dynamic label queue. (a) Precision-recall curve for different iterations. (b) mAP and mAR curve on iterations.

We can observe that the quality of pseudo-labels in the queue seems to saturate after $4k$ iterations from Fig. 1(b), maintaining an mAP50$_{novel}$ around 16.9%.

This is reasonable, as a trivial combination of proposal generator with Remote-CLIP [10] for open-vocabulary detection is suboptimal. However, by employing a hybrid training mechanism, we improve the mAP50$_{novel}$ to 43.3%.

## D  Qualitative Results

### D.1  Visualization of Foreground Score

Fig. 2 illustrates the visualization of RPN foreground confidence score map. While the supervised pipeline initially demonstrates the capability to discover novel category targets, its ability to identify novel categories diminishes as the model becomes adept at accurately object classification and bounding box coordinates regression. The semi-supervised pipeline can only discover categories contained in the labelled data, lacking proficiency in recognizing targets from novel categories. In contrast, Our CastDet progressively enhances its detection capability for novel categories.

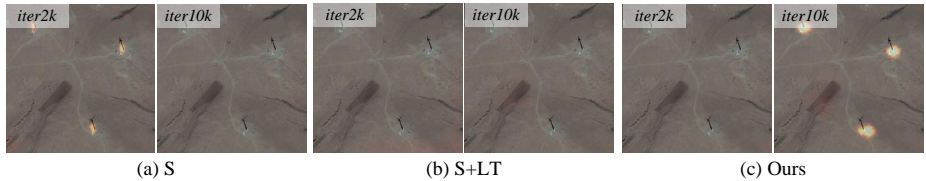

**Fig. 2:** Visualization of RPN foreground confidence during the training process. S, LT, ET denote student, localization teacher and external teacher, respectively.

### D.2  Visualization of Pseudo Labels

In Fig. 3, we show the qualitative results of pseudo-labels under different box selection strategies.

### D.3  Visualization of Predictions.

In Fig. 4 and Fig. 5, we present the qualitative results of CastDet on open-vocabulary COCO and VisDroneZSD, respectively. CastDet demonstrates accurate localization and characterization for both natural and aerial images.

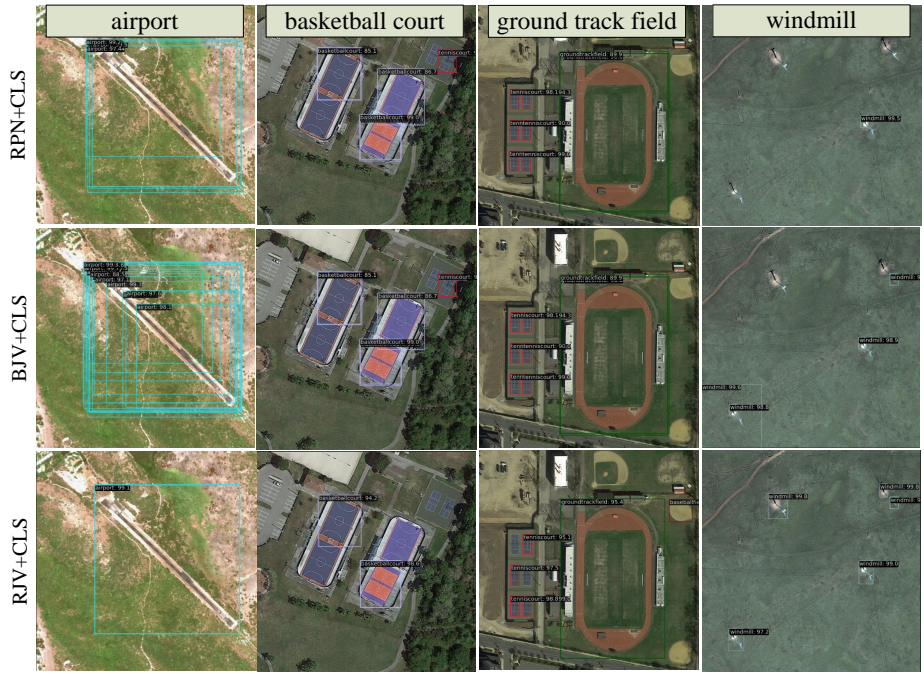

**Fig. 3:** Pseudo labels of different box selection strategies.

# References

1. Bilen, H., Vedaldi, A.: Weakly supervised deep detection networks. In: Proceedings of the IEEE conference on computer vision and pattern recognition. pp. 2846–2854 (2016) 3

2. Cheng, G., Han, J., Lu, X.: Remote sensing image scene classification: Benchmark and state of the art. Proceedings of the IEEE **105**(10), 1865–1883 (2017). https://doi.org/10.1109/JPROC.2017.2675998 1

3. Cheng, G., Zhou, P., Han, J.: Learning rotation-invariant convolutional neural networks for object detection in vhr optical remote sensing images. IEEE Transactions on Geoscience and Remote Sensing **54**(12), 7405–7415 (2016) 1

4. Deng, J., Dong, W., Socher, R., Li, L.J., Li, K., Fei-Fei, L.: Imagenet: A large-scale hierarchical image database. In: 2009 IEEE Conference on Computer Vision and Pattern Recognition. pp. 248–255 (2009). https://doi.org/10.1109/CVPR.2009.5206848 2

5. Feng, C., Zhong, Y., Jie, Z., Chu, X., Ren, H., Wei, X., Xie, W., Ma, L.: Promptdet: Towards open-vocabulary detection using uncurated images. In: European Conference on Computer Vision. pp. 701–717. Springer (2022) 2, 3

6. Gu, X., Lin, T.Y., Kuo, W., Cui, Y.: Open-vocabulary object detection via vision and language knowledge distillation. arXiv preprint arXiv:2104.13921 (2021) 2, 3

7. He, K., Gkioxari, G., Dollár, P., Girshick, R.: Mask r-cnn. In: Proceedings of the IEEE international conference on computer vision. pp. 2961–2969 (2017) 2, 3

**Fig. 4:** Visualization of open-vocabulary COCO inference. Novel categories in this figure include skateboard, elephant, cake, keyboard, cat, tie, airplane, cow, bus, scissors, couch, and dog.

8. Li, K., Wan, G., Cheng, G., Meng, L., Han, J.: Object detection in optical remote sensing images: A survey and a new benchmark. ISPRS journal of photogrammetry and remote sensing **159**, 296–307 (2020) 1

9. Lin, T.Y., Maire, M., Belongie, S., Hays, J., Perona, P., Ramanan, D., Dollár, P., Zitnick, C.L.: Microsoft coco: Common objects in context. In: Computer Vision–ECCV 2014: 13th European Conference, Zurich, Switzerland, September 6-12, 2014, Proceedings, Part V 13. pp. 740–755. Springer (2014) 2

10. Liu, F., Chen, D., Guan, Z., Zhou, X., Zhu, J., Zhou, J.: Remoteclip: A vision language foundation model for remote sensing. arXiv preprint arXiv:2306.11029 (2023) 4

11. team at Lab of Machine Learning, A., Mining, D.: Zero-shot object detection challenge (2023), http://aiskyeye.com/challenge-2023/zero-shot-object-detection/, Last accessed on 2023-11-09 1

12. Xia, G.S., Bai, X., Ding, J., Zhu, Z., Belongie, S., Luo, J., Datcu, M., Pelillo, M., Zhang, L.: Dota: A large-scale dataset for object detection in aerial images. In: 2018 IEEE/CVF Conference on Computer Vision and Pattern Recognition. pp. 3974–3983 (2018). https://doi.org/10.1109/CVPR.2018.00418 1

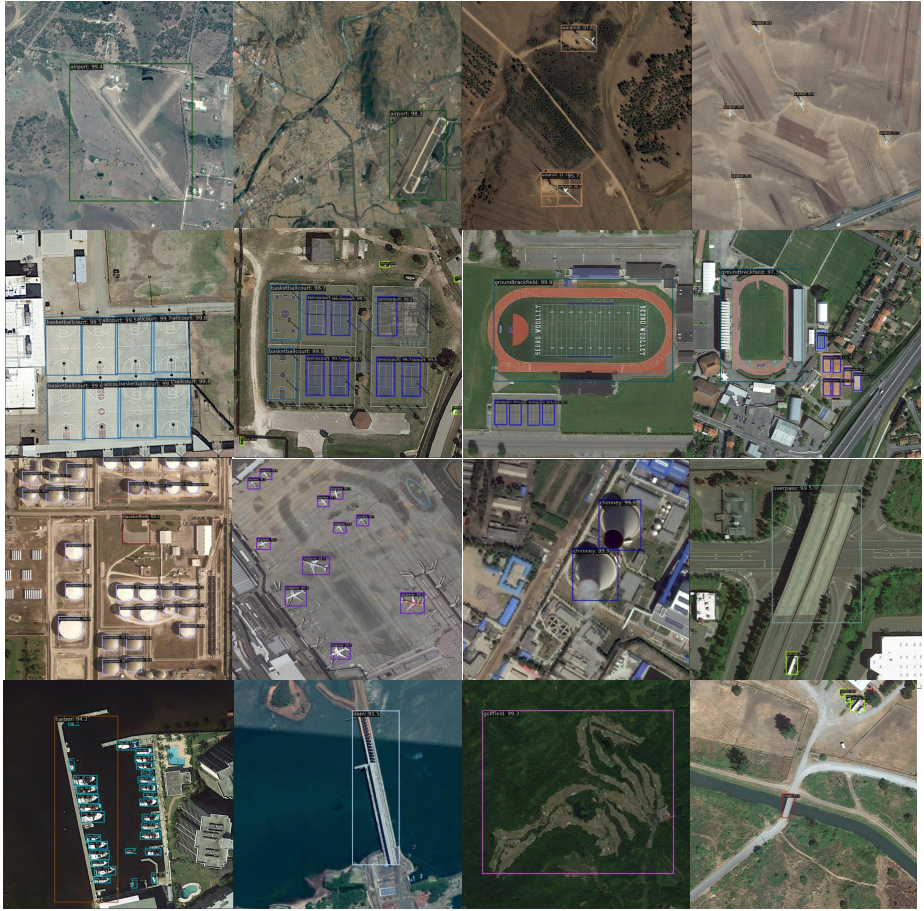

**Fig. 5:** Visualization of open-vocabulary VisDroneZSD inference. Novel categories in this figure include airport, basketball court, ground track field, and windmill.

13. Ye, K., Zhang, M., Kovashka, A., Li, W., Qin, D., Berent, J.: Cap2det: Learning to amplify weak caption supervision for object detection. In: Proceedings of the IEEE/CVF International Conference on Computer Vision. pp. 9686–9695 (2019) 3

14. Zareian, A., Rosa, K.D., Hu, D.H., Chang, S.F.: Open-vocabulary object detection using captions. In: Proceedings of the IEEE/CVF Conference on Computer Vision and Pattern Recognition. pp. 14393–14402 (2021) 2, 3

15. Zhou, X., Girdhar, R., Joulin, A., Krähenbühl, P., Misra, I.: Detecting twenty-thousand classes using image-level supervision. In: European Conference on Computer Vision. pp. 350–368. Springer (2022) 2, 3



\end{document}